\documentclass[letterpaper, 10pt, conference]{ieeeconf}
\IEEEoverridecommandlockouts
\usepackage[english]{babel}
\usepackage{amsmath}
\usepackage{amssymb}
\usepackage{booktabs}
\usepackage{siunitx}
\usepackage{graphicx}
\usepackage{multirow}
\usepackage[inkscapelatex=false]{svg}
\usepackage{amsfonts}
\usepackage{tabularx}
\usepackage{algorithm,algorithmic}
\usepackage{bm}
\usepackage{adjustbox}
\usepackage{xcolor}
\usepackage{xurl}
\usepackage{subcaption}
\usepackage{cite}
\makeatletter
\let\NAT@parse\undefined
\makeatother
\usepackage{hyperref}
\usepackage{relsize}
\usepackage{float}
\usepackage{pifont}
\usepackage{tikz}
\usetikzlibrary{positioning,fit}
\usepackage{diagbox}

\usepackage{capt-of}
\usepackage{comment}
\title{\LARGE \bf ClearNav: Perception-Aligned Quadrotor Flight Using Differential-Physics}

\author{%
  Aditya Azad\textsuperscript{*1},
  Arthur Huang\textsuperscript{*2},
  and Guanrui Li\textsuperscript{1}%
  \thanks{\textsuperscript{*}Equal contributors}
  \thanks{\textsuperscript{1}Authors are with Worcester Polytechnic Institute, MA, USA. {\tt\footnotesize email: aazad@wpi.edu, gli7@wpi.edu}}
  \thanks{\textsuperscript{2}The Author is with Northeastern University, MA, USA. {\tt\footnotesize email: huang.arth@northeastern.edu}}
}%

\begin{document}

\thispagestyle{empty}
\pagestyle{empty}
\makeatletter
\makeatother

\maketitle

\begin{abstract}
Autonomous quadrotor flight in cluttered environment is an key capability for tasks like inspection and search. Navigation policies trained in simulation and transferred to hardware can fly at high speed in open space, but they can still collide as obstacle clearance shrinks. We attribute most of these collisions to a single mechanism, namely motion in directions the onboard camera does not observe. Heading that is not aligned with the direction of travel and lateral excursion during obstacle avoidance both carry the vehicle toward obstacles outside its field of view. We introduce ClearNav, a navigation policy trained with two reward terms that constrain motion to the observed direction, a forward-looking heading target and a penalty on lateral velocity. A third term penalizes speed near obstacles so that an obstacle entering the field of view remains avoidable. In simulation, across 18 held-out scenarios with 2304 paired flights per policy unseen in training, ClearNav improves success by 6.8 points, cuts the collision rate by 65\% relative to the state of the arts, and flies at maximum $5 \si{m/s}$. For the proximity term alone under matched commanded speeds, the collision reduction still holds at 70\%. The effect is significant on static scenes and consistent in direction on moving obstacles, though not significant there at the suite level, showing that safer flight in clutter is possible even on the limited sensors and compute of small drones. We also validate our navigation policy in a real-world cluttered environment, showing that it enables the quadrotor to navigate through the cluttered environment successfully and safely.
\end{abstract}

\section*{Supplementary material}

\textbf{Website}:
\url{https://acp-lab.github.io/clearnav}

\textbf{Code}:
\url{https://github.com/acp-lab/clearnav}

\textbf{Video}:
\url{https://youtu.be/9hldGKQiQgc}

\section{Introduction}\label{sec:introduction}

Autonomous quadrotor flight in cluttered, confined spaces is a core requirement for search and rescue, and for inspecting forests or collapsed structures. Confined geometry restricts platform size, and thus onboard sensing and compute, so the vehicle must stay safe at small clearances under strict computational budgets.

Classical pipelines decompose navigation into mapping, planning, and control~\cite{gaoGradientbasedOnlineSafe2017,zhouRobustEfficientQuadrotor2019,liuSearchbasedMotionPlanning2017a,zhouRAPTORRobustPerceptionaware2020,xuRealtime3DNavigation2015,richterPolynomialTrajectoryPlanning2016,liuSearchbasedMotionPlanning2017,daiFastFrontierbasedInformationdriven2020}. Inaccuracies propagate across stages, and the accumulated latency is too high for fast flight in clutter. Hybrid learning-plus-optimization variants recover some latency but still rely on an explicit map~\cite{tangLearningTrajectoriesReal2018,wuDeepLearningOptimization2023,jiangSelfSupervisedLearningApproach2025}, as do pure optimization methods~\cite{kongAvoidingDynamicSmall2021}. Their fixed-rate replanning leaves less margin as speed grows, and all demand dense sensing with a heavy SWaP burden on small platforms. Imitation learning replaces the pipeline with a direct sensorimotor mapping~\cite{loquercioLearningHighSpeedFlight2021,xingBootstrappingReinforcementLearning2024,lowSOUSVIDECooking2025}, but the student is bounded by the expert's performance and degrades out of distribution when the expert uses privileged information~\cite{xingBootstrappingReinforcementLearning2024}. Reinforcement learning removes the expert~\cite{songAutonomousDroneRacing2021,kaufmannChampionlevelDroneRacing2023,romeroDreamFlyModelBased2025,hanReactiveAerobaticFlight2025,gelesDemonstratingAgileFlight2024,malczykReinforcementLearningActive2026,zhangHighspeedVisionbasedFlight2026}, but its sampled gradients need thousands of parallel environments. Model-based variants trade sample efficiency for a world model whose long-horizon rollouts can fail to generalize~\cite{verraestSkyDreamerInterpretableEndtoEnd2025,vytlaDreamingFalconPhysicsInformed2025}, and reward shaping reaches the policy only through indirect, high-variance sampled correlations.
\begin{figure}
    \vspace{9pt}
    \centering
    \hspace*{0.15\columnwidth}%
    \includesvg[width=0.85\columnwidth]{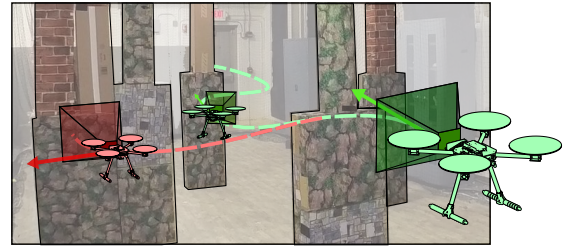}%
    \caption{ClearNav, an end-to-end navigation policy trained through differentiable physics, enables a drone to fly through cluttered environment while dodging obstacles by aligning its camera with its direction of travel and slowing down near obstacles.}
    \label{fig:placeholder}
\end{figure}

Differentiable physics instead differentiates the entire rollout and updates the policy by backpropagation through the simulator~\cite{leeQuadrotorNavigationUsing2026,heegLearningQuadrotorControl2025,zhangLearningVisionbasedAgile2025,suVectorFieldAugmented2026,zhongPanoDPLearningCollisionFree2026,liSimpleStableFast2026,wangLearnVariationVariationally2026}, an idea that extends to perceptive legged control~\cite{luoResidualPolicyLearning2025}. Some methods retain an RL actor or critic and are better read as first-order hybrids~\cite{chenGRaDNavEfficientlyLearning2025,xing2024stabilizing}, surveyed in~\cite{newburyReviewDifferentiableSimulators2024}. The analytic gradients generalize to new environments from few simulation samples~\cite{zhangLearningVisionbasedAgile2025}, but differentiating long unrolled rollouts is ill-conditioned and the dynamics Jacobian can become chaotic~\cite{metzGradientsAreNot2022}. The point-mass dynamics sacrifice rotational degrees of freedom, and the privileged shaping signals are unavailable at deployment. As a result, these policies still struggle in environments that need deliberate, multi-step avoidance rather than purely reactive maneuvers: the physics-based rewards that keep training tractable give little incentive for such behavior. Later work shapes the objective with global information available at training time~\cite{leeQuadrotorNavigationUsing2026,suVectorFieldAugmented2026}, but this oracle is not used to its full potential. In our experiments these approaches fail in cluttered environments, with two recurring causes: misaligned yaw during motion, and overly wide, unsafe trajectories. The point-mass lineage~\cite{zhangLearningVisionbasedAgile2025} penalizes only the closing-speed component of velocity, leaving tangential motion free. General-purpose GPU frameworks such as DiffAero~\cite{zhangDiffAeroGPUAcceleratedDifferentiable2025} accelerate training of such policies but are validated on agile-flight benchmarks rather than deliberate avoidance in tight clutter. We introduce a direction-agnostic speed-proximity cost as one term in a navigation objective for tight clutter, paired with heading and strafe-aware penalties.

To address this gap, we introduce \textbf{ClearNav}, a safer differentiable-physics policy for cluttered spaces. ClearNav adds three reward terms to an already-tuned differentiable-rollout objective~\cite{leeQuadrotorNavigationUsing2026}, chosen so that no degree of freedom is sacrificed.

Specifically, the contributions of this work are:

\begin{itemize}
\item \textbf{Safer navigation in cluttered environments}: We penalize high speed near obstacles, misaligned yaw, and excessive strafing. Each penalized behavior is observable at inference time, so the policy keeps the information it needs to act on the objective during deployment.
\item \textbf{Navigation around large obstacles}: Our policy navigates around obstacles across a wide range of sizes, shapes, and counts across simulations and real world experiment including large obstacles.
\item \textbf{Thorough simulation experiments}: We compare against other exisiting approaches on static out-of-distribution scenes, a matched-velocity control, moving obstacles, and a cross-framework setting.
\end{itemize}

\providecommand{\bm}[1]{\boldsymbol{#1}}
\providecommand{\Fw}{\mathcal{W}}
\providecommand{\Fs}{\mathcal{S}}
\providecommand{\Fb}{\mathcal{B}}
\providecommand{\Loa}{\mathcal{L}_{\mathrm{oa}}}

\section{Methodology}

This section describes how the policy is trained. Section~\ref{sec:problem} states the navigation problem and defines the observation and action. Section~\ref{sec:dynamics} introduces the differentiable point-mass dynamics that let us backpropagate through entire rollouts. Section~\ref{sec:env} describes the Habitat training scenes and the geodesic guidance field that shapes the objective without entering the policy. Section~\ref{sec:arch} presents the three-branch recurrent policy architecture. Section~\ref{sec:objective} details the per-step objective and the three loss terms we add to it. Section~\ref{sec:training} describes the two-stage training procedure that carries the weights from an empty arena into cluttered scenes.

\subsection{Problem formulation}
\label{sec:problem}

\begin{figure}[!t]
  \vspace*{3pt}
  \centering
  \includesvg[width=\linewidth]{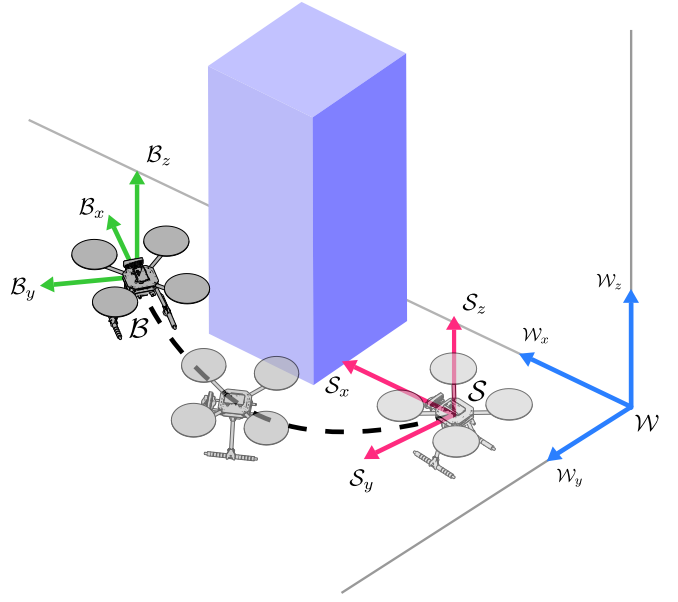}
  \caption{Frames used in our formulation. All frames use right handed convention with positive z-axis pointing opposite to gravity at hover. World frame is indicated with blue, start frame with red and body frame with green.}
  \label{fig:frames}
\end{figure}

\begin{figure*}[!t]
  \vspace*{3pt}
  \centering
  \includesvg[width=0.8\linewidth]{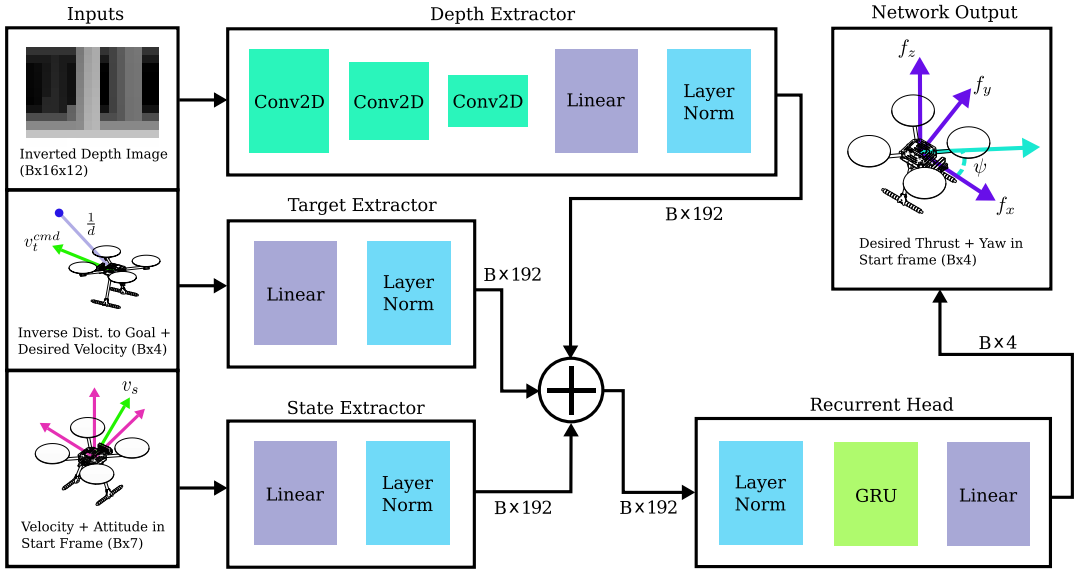}
  \vspace{1pt}
  \caption{Policy architecture. The three modality branches are projected to a
  common width $n_f$ and summed, so the recurrent input dimension does not grow
  with the number of sensors.}
  \label{fig:arch}
\end{figure*}

A quadrotor flies autonomously from a randomly sampled start pose to a goal position $\bm{p}_g \in \mathbb{R}^3$ through an unknown, cluttered environment, using only a forward-facing depth camera and proprioception. We use three frames (Fig.~\ref{fig:frames}): an ENU world frame $\Fw$, a start frame $\Fs$ fixed at the reset pose, and a body frame $\Fb$ at the center of mass, with rotations $R_{ws}, R_{wb}\in SO(3)$. At each step $t$ the policy receives the observation
\begin{equation}
  \bm{o}_t \;=\; \bigl(\, \bm{s}_t,\; \bm{g}_t,\; D_t \,\bigr),
  \label{eq:obs}
\end{equation}
where the proprioceptive state $\bm{s}_t \in \mathbb{R}^{7}$ stacks the attitude quaternion of $\Fb$ relative to $\Fs$ and the velocity expressed in $\Fs$, both perturbed by Gaussian noise resampled every step. The goal encoding $\bm{g}_t \in \mathbb{R}^{4}$ is a clamped proportional velocity command $\bm{v}^{\mathrm{cmd}}_t = \min(K_p\|\bm{p}_g - \bm{p}_t\|,\, v^\star)\,\widehat{(\bm{p}_g - \bm{p}_t)}$, where $K_p$ is a proportional gain and $v^\star$ is the command speed cap, together with an inverse range $[\max(\|\bm{p}_g - \bm{p}_t\|,\, d_{\min})]^{-1}$, with $d_{\min} = 0.5$\,m, and the hat $\widehat{(\,\cdot\,)}$ denotes the normalized unit vector of a vector. The command points straight at the goal and carries no obstacle information, so all avoidance must come from the depth image $D_t \in \mathbb{R}^{1 \times H \times W}$ of height $H$ and width $W$, whose invalid returns are set to the far plane and treated as free space.

The action is a mass-normalized collective thrust vector $\bm{f}_t$ in $\Fs$ plus an absolute yaw setpoint $\psi_t$,
\begin{equation}
  \bm{a}_t = \bigl(\bm{f}_t,\; \psi_t\bigr) \in \mathbb{R}^3 \times (-\pi, \pi].
  \label{eq:action}
\end{equation}
We denote our navigation policy as $\pi_\theta$, parameterized by $\theta$. The navigation policy will take observation $\bm{o}_t$ and the policy's recurrent state $\bm{h}_t$ as input and output control action $\bm{a}_t$, so $\bm{a}_t = \pi_\theta(\bm{o}_t, \bm{h}_t)$.
Since the simulator is differentiable end to end, we optimize $\theta$ directly rather than via policy gradients:
\begin{equation}
  \min_{\theta}\;
  \mathbb{E}\Bigl[\, \tfrac{1}{H}\!\sum_{t=0}^{H-1} \gamma^{t}\,
  \mathcal{L}\bigl(\bm{x}_t, \bm{a}_t\bigr) \Bigr],
  \label{eq:problem}
\end{equation}
where $\bm{x}_t$ is the full simulator state, $H$ the rollout horizon, and the expectation is over initial conditions, scenes, and the injected noise. The gradient $\partial \mathcal{L} / \partial \theta$ is obtained by backpropagation through the unrolled dynamics (Sec.~\ref{sec:training}).

\subsection{Differentiable dynamics}
\label{sec:dynamics}

The vehicle is modeled as a point mass with an attitude reconstructed from the commanded thrust and yaw~\cite{leeQuadrotorNavigationUsing2026, zhangLearningVisionbasedAgile2025}, an abstraction that can struggle where significant yaw motion is required. Translational dynamics are integrated with velocity Verlet at step $\Delta t$ under net acceleration $\bm{a}^{\mathrm{net}}_t = R_{ws}\bm{f}_t + \bm{g}$,
\begin{align}
  \bm{p}_{t+1} &= \bm{p}_t + \bm{v}_t \Delta t
                  + \tfrac{1}{2}\bm{a}^{\mathrm{net}}_t \Delta t^2, \\
  \bm{v}_{t+1} &= \bm{v}_t
                  + \tfrac{1}{2}\bigl(\bm{a}^{\mathrm{net}}_t
                  + \bm{a}^{\mathrm{net}}_{t+1}\bigr)\Delta t .
  \label{eq:verlet}
\end{align}
Gravity $\bm{g}$ is resampled per episode from $\mathcal{N}(-9.81\,\bm{e}_3, \sigma_g^2)$ on the vertical axis, where $\bm{e}_3$ is the world vertical unit vector and $\sigma_g$ is the gravity noise standard deviation, which absorbs unmodeled mass and thrust-coefficient error.

Attitude is not integrated. Given the filtered thrust command $\bm{f}$ and yaw setpoint $\psi$, the body frame is reconstructed in closed form from its axes $\bm{x}_b, \bm{y}_b, \bm{z}_b$ so that $\bm{z}_b$ is along the thrust and $\bm{x}_b$ is as close as possible to the commanded heading $\bm{h} = R_{ws}R_z(\psi)\bm{e}_1$, where $R_z(\psi)$ rotates about the vertical axis by $\psi$ and $\bm{e}_1$ is the start-frame forward axis, with $\bm{y}_b$ completing the right-handed orthonormal triad,
\begin{equation}
  \bm{z}_b = \widehat{\bm{f}}, \quad
  \bm{y}_b = \widehat{\bm{z}_b \times \bm{h}}, \quad
  \bm{x}_b = \widehat{\bm{y}_b \times \bm{z}_b},
  \label{eq:attitude}
\end{equation}
with degenerate cases handled by fixed fallbacks. Body rate $\bm{\omega}$, recovered by finite-differencing consecutive rotations, is differentiable and penalized directly.

Unrolling \eqref{eq:verlet} for hundreds of steps produces the same ill-conditioning that afflicts long BPTT chains. To solve this, gradients flowing back through the BPTT rollout are shrunk by $e^{-\lambda_d T}$ over a horizon $T$, bounding the total gain.

\subsection{Environment and geodesic guidance}
\label{sec:env}

Training scenes are built on Habitat-sim~\cite{habitat}. The nearest-obstacle query is exact: the closest surface point $\bm{c}_t$ gives a signed clearance
\begin{equation}
  d_t \;=\; \|\bm{c}_t - \bm{p}_t\| - r ,
  \label{eq:clearance}
\end{equation}
where $r$ is the vehicle's collision radius. An episode terminates in collision when $d_t < 0$ or on leaving the stage, and in success when $\|\bm{p}_g - \bm{p}_t\| < r_{\mathrm{succ}}$.

\begin{equation}
  \widehat{\bm{u}}(\bm{p}) \;=\; -\widehat{\nabla \Psi(\bm{p})}
  \label{eq:geodesic}
\end{equation}
where $\Psi$ is the scalar time-of-arrival potential and $\widehat{\bm{u}}$ is its unit descent direction. This field is privileged, only shaping the objective and never an input to $\pi_\theta$.

\subsection{Policy architecture}
\label{sec:arch}

$\pi_\theta$ is a three-branch encoder followed by a single recurrent cell and a linear head (Fig.~\ref{fig:arch}). The depth branch inverts the range map, $\tilde{D} = 1/(D + \epsilon)$, then max-pools it to a coarse $H' \times W'$ grid. Max-pooling inverse depth preserves the closest obstacle in each cell. Three stride-1 convolutions with $32/64/128$ channels and leaky-ReLU, a linear projection to $n_f$, and layer normalization follow. The state and goal branches are each a layer-normalized MLP to width $n_f$.

The three feature vectors are summed, layer-normalized, passed through a GRU cell of width $n_f$, and mapped by one linear layer to an unbounded thrust vector and a yaw angle squashed by a sigmoid into $(-\pi,\pi]$. The whole network is $\approx\!2.2$M parameters, which can run on many low power single board computers.

\subsection{Objective}
\label{sec:objective}

\begin{figure*}[!t]
  \vspace*{3pt}
  \centering
  \includesvg[width=0.8\linewidth]{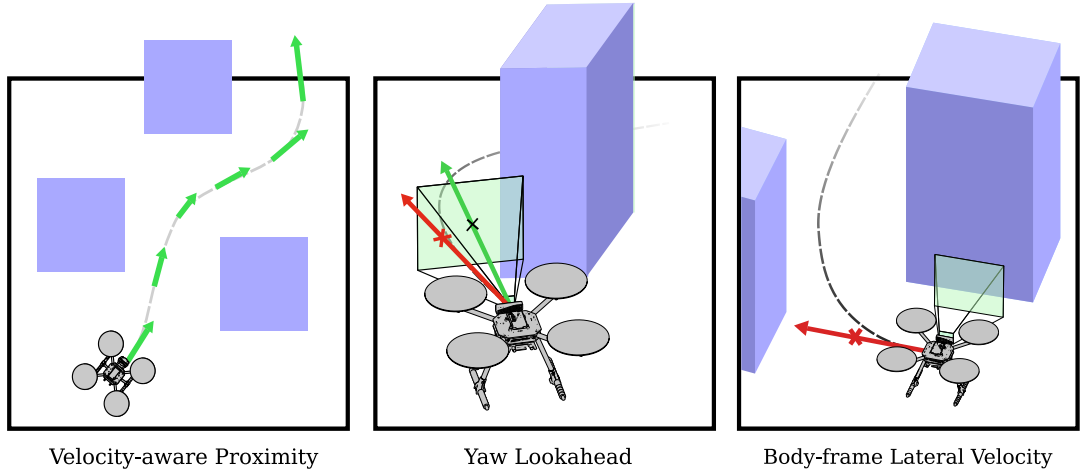}
  \vspace{1pt}
  \caption{Our modifications to the loss function include three components: a velocity-aware proximity penalty, which penalizes high speed near obstacles regardless of direction; a yaw lookahead reward, which encourages the policy to align its heading with the anticipated velocity direction; and a body-frame lateral-velocity penalty, which discourages sideways motion.}
  \label{fig:losses}
\end{figure*}

The per-step cost is a weighted sum of nine terms. Writing $\bar{\bm{v}}$ for the uniform moving average of velocity over a window $T_{\mathrm{avg}}$, $\bar{\bm{v}}^{\,e}$ for its exponentially weighted counterpart, $\bm{v}_b = R_{wb}^{\!\top}\bm{v}$ for the velocity expressed in $\Fb$, and $\widehat{\bm{c}}_t$ for the unit vector toward the nearest obstacle,
\begin{equation}
\begin{aligned}
  \mathcal{L} \;=\;\; & \lambda_v \mathcal{L}_v
  + \lambda_{\bar v} \mathcal{L}_{\bar v}
  + \lambda_c \mathcal{L}_c
  + \!\!\!\sum_{k \in \{a, j, \omega\}}\!\!\! \lambda_k \mathcal{L}_k \\[2pt]
  & {}+ \lambda_{\mathrm{oa}} \Loa
  + \lambda_{\mathrm{slip}} \mathcal{L}_{\mathrm{slip}}
  + \lambda_\psi \mathcal{L}_\psi \\[2pt]
\end{aligned}
  \label{eq:total}
\end{equation}
and the reward reported in the logs is $r = -\mathcal{L}$.

\subsubsection{Progress}
The vehicle is asked to track the routed direction at the commanded speed, penalized with a Huber loss on the magnitude of the tracking error, plus a one-sided quadratic that discourages exceeding the commanded speed:
\begin{align}
  \mathcal{L}_v      &= \mathrm{huber}\bigl(\, \|v^\star \widehat{\bm{u}}
                          - \bar{\bm{v}}\| \,\bigr),
  \label{eq:lv}\\
  \mathcal{L}_{\bar v} &= \bigl(\|\bm{v}\| - v^\star\bigr)_+^2 .
  \label{eq:lvmax}
\end{align}
Using the windowed average $\bar{\bm{v}}$ rather than the instantaneous velocity in \eqref{eq:lv} is what allows a lateral excursion around an obstacle without being immediately penalized for it.

\subsubsection{Collision}
The safety term combines a closing-speed penalty with a soft interior barrier,
\begin{equation}
  \mathcal{L}_c =
    \underbrace{\bigl(\bm{v}\!\cdot\!\widehat{\bm{c}}\bigr)_+
                \bigl(d_f - d\bigr)_+^2}_{\text{closing speed}}
  \;+\;
    \underbrace{\beta_1 \log\!\bigl(1 + e^{\beta_2 d}\bigr)}_{\text{barrier}} ,
  \label{eq:lc}
\end{equation}
Here $d$ is the clearance $d_t$ from \eqref{eq:clearance}, and $\beta_2 < 0$ so the barrier is negligible for $d \gg 0$ and grows linearly once the surface is penetrated. The closing-speed factor penalizes radial approach within a shell of thickness $d_f$.

\subsubsection{Smoothness}
$\mathcal{L}_a = \|\bm{a}\|^2$, $\mathcal{L}_j = \|\bm{j}\|^2$ and $\mathcal{L}_\omega = \|\bm{\omega}\|^2$ penalize acceleration, jerk, and body rate, where here $\bm{a}$ is the translational acceleration and $\bm{j}$ the jerk, making trajectories flyable.

\subsubsection{Heading}
The heading reference interpolates along the great circle from the goal direction to the smoothed velocity direction, since early in an episode there is no motion to look along,
\begin{equation}
  \widehat{\bm{y}}^{\,\mathrm{ref}}
    = \mathrm{slerp}\bigl(\, \widehat{\bm{p}_g - \bm{p}},\;
        \widehat{\bar{\bm{v}}^{\,e}};\; \mathrm{clip}(\|\bm{p} - \bm{p}_0\|, 0, 1) \,\bigr),
  \label{eq:yawref}
\end{equation}
\begin{equation}
  \mathcal{L}_\psi = -\,\widehat{\bm{y}}^{\,\mathrm{ref}} \!\cdot\! \widehat{\bm{y}},
  \label{eq:lyaw}
\end{equation}
where $\widehat{\bm{y}}$ is the body $x$ axis projected onto the horizontal plane, blended by distance traveled since the reset position $\bm{p}_0$, clamped to $[0,1]$. $\mathcal{L}_\psi$ is gated off within $1\,\mathrm{m}$ of the goal, where the goal direction is ill-conditioned.

When geodesic guidance is available, the slerp reference of \eqref{eq:yawref} is replaced by the routed direction sampled $\tau_\ell$ seconds ahead of the vehicle along its smoothed velocity,
\begin{equation}
  \boxed{\;\widehat{\bm{y}}^{\,\mathrm{ref}} \;=\; -\,\widehat{\nabla \Psi\bigl(\bm{p} + \tau_\ell\,\bar{\bm{v}}^{\,e}\bigr)}\;},
  \label{eq:yawla}
\end{equation}
wherever that gradient is valid. Where the look-ahead point lies inside geometry, the slerp fallback of \eqref{eq:yawref} is kept. Sampling ahead of $\bm{p}$ turns the heading before a bend, so the depth camera leads the path. This gradient, like $\widehat{\bm{u}}$ in \eqref{eq:lv}, only shapes training and is never an input to $\pi_\theta$.

\subsubsection{Velocity-aware proximity cost}
\label{sec:newloss}

Equation~\eqref{eq:lc} penalizes only radial approach, so any velocity orthogonal to $\widehat{\bm{c}}$ is free, and its barrier fires only once $d$ is near zero, too late for a vehicle whose stopping distance at 5\,m/s is several meters~\cite{zhangLearningVisionbasedAgile2025}. We therefore add a cost on \emph{being fast while close}:
\begin{equation}
  \boxed{\;\Loa \;=\; \|\bm{v}\|\, \exp\!\bigl(-2\,(d - \delta)_+\bigr)\;}
  \label{eq:loa}
\end{equation}
where $\delta$ is an inflation radius.

The term is direction-agnostic, since $\Loa$ depends only on $\|\bm{v}\|$, so the only way to reduce it near an obstacle is to slow down. Its gradient $\partial \Loa / \partial \bm{v} = \widehat{\bm{v}}\exp(-2(d-\delta)_+)$ opposes the current velocity with magnitude set by proximity, and the exponential falloff keeps $\Loa$ negligible in open space, so it can be added without re-tuning the other weights.

\subsubsection{Body-frame lateral-velocity penalty}
\eqref{eq:lyaw} asks the camera to look where the vehicle is going, but thrust is commanded in $\Fs$, not $\Fb$, so lateral body velocity is no costlier than forward and a policy can satisfy $\mathcal{L}_\psi$ while flying broadside. We close this with a penalty on the body-frame lateral velocity,
\begin{equation}
  \boxed{\;\mathcal{L}_{\mathrm{slip}} \;=\; v_{\mathrm{lat}}^{\,2}\,\bigl(\|\bm{v}\| + 1\bigr),
    \qquad v_{\mathrm{lat}} = \bigl(\bm{v}_b\bigr)_y\;},
  \label{eq:lslip}
\end{equation}
where $v_{\mathrm{lat}}$ is the strafe-axis component of $\bm{v}_b = R_{wb}^{\!\top}\bm{v}$. The speed factor $\|\bm{v}\| + 1$ is detached from the autograd graph, so gradient flows only through the quadratic on $v_{\mathrm{lat}}$, making fast sideways motion expensive while leaving the small transient lateral component of a coordinated turn nearly free. $\mathcal{L}_{\mathrm{slip}}$ carries no obstacle or scan reference, so unlike $\Loa$ it cannot be farmed by routing around geometry, and its gradient is a pure anti-strafe command leaving the forward and vertical components free.

\subsubsection{Training}
\label{sec:training}

We roll $N$ agents forward for $H$ steps, accumulate the discounted cost of \eqref{eq:problem}, and backpropagate through the whole unrolled graph in a single backward pass, covering the dynamics, the clearance query, and the reward. Parameters are updated with AdamW under a cosine-annealed learning rate and gradient-norm clipping at $g_{\max}$, and after each iteration the simulator and recurrent states are detached, truncating the graph while preserving physical continuity. Agents that terminate are reset individually, with the discount and recurrent state zeroed for those rows only.

Training proceeds in two stages that carry the weights forward and continue the global iteration counter. The first stage uses an empty walled arena so the policy learns the thrust-to-motion mapping and goal reaching, and the second stage uses cluttered scenes with time-of-arrival guidance and the full objective.

\providecommand{\Loa}{\mathcal{L}_{\mathrm{oa}}}

\section{Experimental Results}\label{sec:experiments}

We validate our methods by answering four questions: 
\begin{enumerate}
  \renewcommand\labelenumi{\textbf{Q\theenumi}}
  \item Does ClearNav help on the scenes it was trained on (Sec.~\ref{sec:indist})?
  \item Does it help out of distribution, on obstacle configurations, densities, arena sizes, and speed bands never seen in training (Sec.~\ref{sec:staticsuite})?
  \item Does the benefit survive when the confounding factors are controlled (Sec.~\ref{sec:matched})?
  \item Does the benefit survive when the obstacles move (Sec.~\ref{sec:moving})?
\end{enumerate}

The first is answered in the negative on the training distribution, where the proximity term adds nothing measurable (Sec.~\ref{sec:indist}). The bundle's gains appear only out of distribution (Sec.~\ref{sec:staticsuite}). A few static scenarios where the baseline matches or beats it, and a weaker, non-significant effect on moving obstacles (Sec.~\ref{sec:moving}), bound the scope of the claim.

\begin{figure}[!t]
  \vspace*{3pt}
  \centering
  \includegraphics[width=\columnwidth]{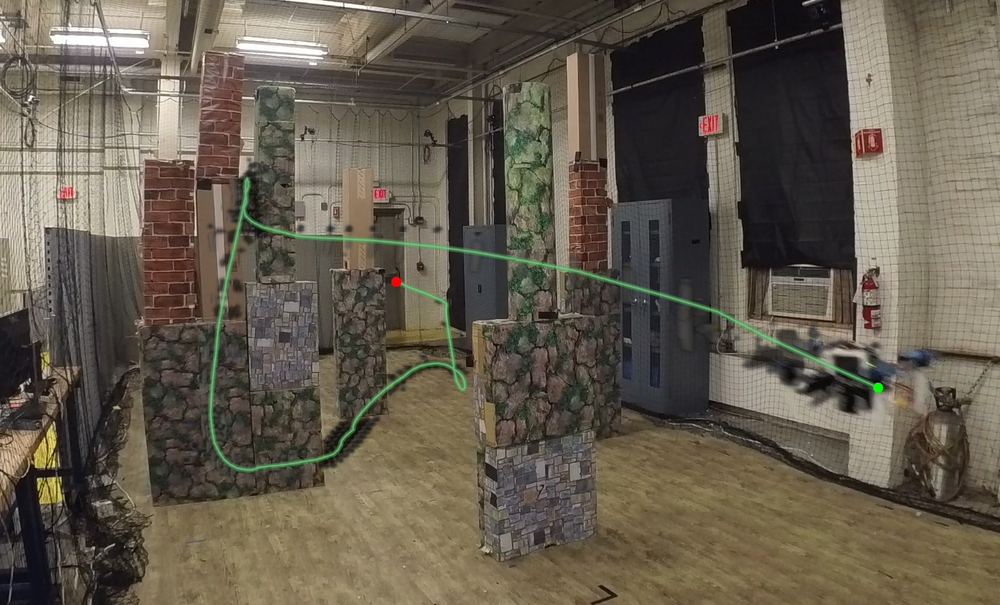}
  \caption{Real world trajectory flown by the ClearNav policy. The drone moves left to avoid the first obstacle, then after looking at it tries to go below, yaws to the right to avoid the second, wider obstacle, and continues to the goal.}
  \label{fig:realworld}
\end{figure}

\begin{figure}[!t]
  \vspace*{3pt}
  \centering
  \includegraphics[width=\columnwidth]{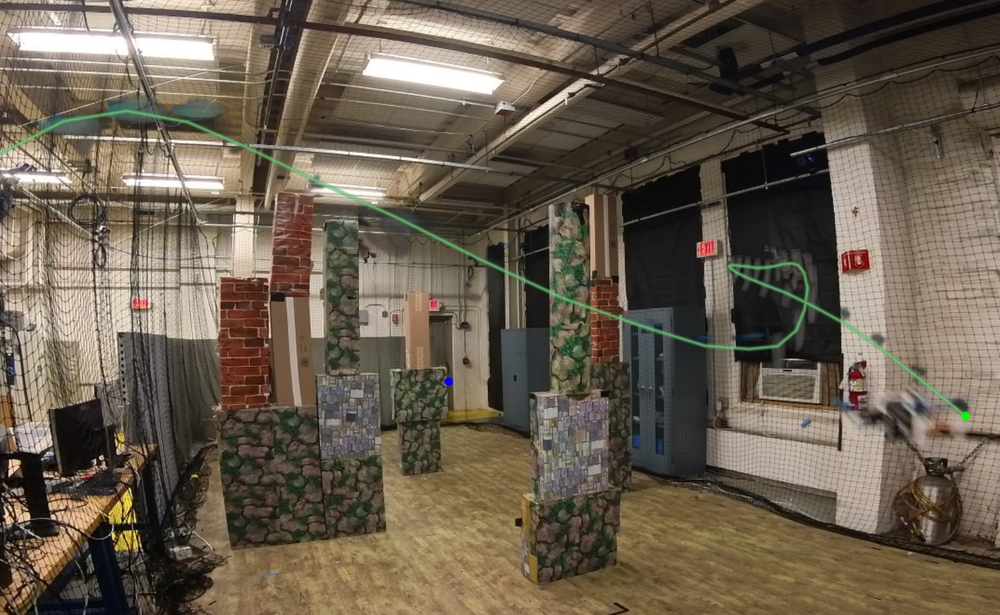}
  \caption{Real world trajectory flown by the baseline policy. The drone drifts toward the net to the side of the course, crashes into it, and never reaches the goal.}
  \label{fig:realworld-baseline}
\end{figure}

\subsection{Experiment Setup}
\label{sec:setup}

We compare four policies:
\begin{itemize}
  \item \textbf{B} (baseline): the original DepthNav reward, i.e.\ \eqref{eq:total} with the three new terms disabled (avoidance driven by $\mathcal{L}_c$ alone); stage-1 cruise speed $\sim\mathcal{U}[1,5]$~m/s.
  \item \textbf{ClearNav} (ours): the full objective activating $\Loa$, $\mathcal{L}_{\mathrm{slip}}$, and the look-ahead yaw of \eqref{eq:yawla}; cruise speed $\sim\mathcal{U}[0.5,10]$~m/s, with a longer stage-1 episode limit for the slower end.
  \item \textbf{OA-only} (ablation): $\Loa$ alone on top of B; appears only in the matched-velocity control (Sec.~\ref{sec:matched}) and speed sweep (Sec.~\ref{sec:moving}), where the full bundle was not re-run.
  \item \textbf{DA} (DiffAero reference): the v1 policy trained in the DiffAero simulator~\cite{zhangDiffAeroGPUAcceleratedDifferentiable2025}, evaluated in that simulator on the same scenario specifications. Sec.~\ref{sec:crossframework} explains why this benchmarks whole stacks rather than the reward term.
\end{itemize}

\noindent
\emph{This is not a single-variable ablation.} ClearNav differs from B in three objective terms ($\Loa$, $\mathcal{L}_{\mathrm{slip}}$, and the look-ahead yaw reference) and the commanded-speed distribution used during stage-1 training. The latter directly causes the mean-speed gap in every table below. Sec.~\ref{sec:matched} isolates it with a control experiment. OA-only provides one ablation point, the proximity term alone. Separating $\mathcal{L}_{\mathrm{slip}}$ from the look-ahead yaw reference is not attempted here.

All evaluations use held-out scenes, with seeds disjoint from training and no shared scene, layout, or start--goal pair. Both policies run under the same environment seed, so poses, goals, and obstacle placements (and trajectories, Sec.~\ref{sec:moving}) are \emph{paired}. An episode is a success within $r_{\mathrm{succ}}=0.5$\,m of the goal, a collision on geometry contact or leaving the stage, or a timeout otherwise. The collision radius is $0.25$\,m, below the $0.3$\,m used in training. Statistical statements are Wilcoxon signed-rank tests over \emph{scenarios}, with each scenario's paired success rate as one matched observation. We include footage of the simulation environments in the supplementary video for the reader.

\providecommand{\Loa}{\mathcal{L}_{\mathrm{oa}}}

\subsection{In-distribution evaluation}
\label{sec:indist}

During training, baseline and OA-only are indistinguishable on a held-out set from the stage-1 scene family, sampled every 100 iterations at 80 episodes and commanded $[1,5]$~m/s. Both converge to $\approx\!99\%$ success and $\approx\!1\%$ collision, with identical mean speed $\approx\!2.5$~m/s. On the training distribution the proximity term thus buys nothing measurable, and costs nothing: it does not slow the policy, degrade goal-reaching, or destabilize training. Whether the bundle matters must be decided out of distribution.

\subsection{Static generalization suite}
\label{sec:staticsuite}

To probe generalization we built an 18-scenario suite sweeping obstacle count $|\mathcal{O}| \in [5, 80]$, shape mix, arena size $L \in [10, 25]$~m, commanded speed band $[0.5,1.5]$ to $[7,10]$~m/s, and spatial clustering. Obstacles form a 2D Gaussian blob centered on the goal, so every path must traverse the clutter. The goal sits at the arena center and the vehicle spawns on a cylinder of radius $L/2$. Each scenario is 16 scenes $\times$ 8 rollouts = 128 paired episodes per policy, 2304 total, with a 1500-step, 30~s limit.

Table~\ref{tab:static} shows the innovations raise mean success from $90.3\%$ to
$97.1\%$ while cutting the collision rate from $8.2\%$ to $2.9\%$, a $65\%$
relative reduction. The improvement is consistent, not driven by a few scenarios. ClearNav is better on 17 of 18 scenarios and worse on none, with one exact tie. This is significant under a Wilcoxon signed-rank test over scenarios, Holm-corrected $p = 3\times10^{-4}$. Pooling episodes gives $2237/2304$ against $2080/2304$.

The gain is largest where \eqref{eq:loa} predicts: dense clutter, clustering, and the tight arena. In these scenes, the baseline loses most, while the near-saturated very-sparse scenario shows no difference. The tight-cluster case is ClearNav's weakest, with a higher collision rate, plausibly dominated by its higher speed.

This is achieved while flying faster, not slower. ClearNav's mean speed of $3.39$~m/s against B's $2.24$~m/s is a $52\%$ increase, the opposite of the usual safety--speed trade-off. Most of the speed gap comes from the wider stage-1 training band rather than the objective terms. Whether the safety gain is attributable to the objective terms or the training band is resolved by Sec.~\ref{sec:matched}.

\begin{table*}[t]
\vspace*{3pt}
\centering
\caption{Static generalization suite: 18 scenarios, 128 episodes per scenario per
policy. \emph{DA}~=~DiffAero~v1 (cross-simulator; Sec.~\ref{sec:crossframework}), \emph{B}~=~DepthNav baseline, \emph{ClearNav}~=~ours (baseline$+\;\mathcal{L}_{\mathrm{oa}} +\;\mathcal{L}_{\mathrm{slip}} +$ look-ahead yaw of \eqref{eq:yawla}). Best of B/ClearNav in bold.}
\label{tab:static}
\footnotesize
\setlength{\tabcolsep}{2.5pt}
\renewcommand{\arraystretch}{0.85}
\begin{tabular}{l rrc rrr rrr rr}
\toprule
& & & & \multicolumn{3}{c}{Success rate (\%)} & \multicolumn{3}{c}{Collision rate (\%)} & \multicolumn{2}{c}{Mean speed (m/s)} \\
\cmidrule(lr){5-7}\cmidrule(lr){8-10}\cmidrule(lr){11-12}
Scenario & $|\mathcal{O}|$ & $L$ & $v^\star$ band & DA & B & ClearNav & DA & B & ClearNav & B & ClearNav \\
\midrule
Baseline & 30 & 25 & 3.0--6.0 & 75.4 & 89.8 & \textbf{96.1} & 12.7 & 10.2 & \textbf{3.9} & 2.36 & \textbf{3.53} \\
Sparse & 10 & 25 & 3.0--6.0 & 90.6 & 94.5 & \textbf{99.2} & 2.3 & 5.5 & \textbf{0.8} & 2.40 & \textbf{3.59} \\
Dense & 60 & 25 & 3.0--6.0 & 53.6 & 87.5 & \textbf{95.3} & 24.9 & 11.7 & \textbf{4.7} & 2.33 & \textbf{3.46} \\
Spheres Heavy & 30 & 25 & 3.0--6.0 & 73.8 & 90.6 & \textbf{96.9} & 13.1 & 8.6 & \textbf{3.1} & 2.37 & \textbf{3.51} \\
Cubes Heavy & 30 & 25 & 3.0--6.0 & 82.1 & 91.4 & \textbf{99.2} & 6.9 & 8.6 & \textbf{0.8} & 2.36 & \textbf{3.56} \\
Short Dist & 20 & 15 & 3.0--6.0 & 66.4 & 89.1 & \textbf{99.2} & 17.0 & 10.9 & \textbf{0.8} & 2.12 & \textbf{3.39} \\
Slow Targets & 30 & 25 & 1.5--3.0 & 57.7 & 87.5 & \textbf{99.2} & 4.0 & 7.8 & \textbf{0.8} & 1.73 & \textbf{2.10} \\
Fast Targets & 30 & 25 & 5.0--8.0 & 56.1 & 90.6 & \textbf{96.9} & 39.8 & 9.4 & \textbf{3.1} & 2.55 & \textbf{4.42} \\
Clustered & 40 & 25 & 3.0--6.0 & 51.1 & 87.5 & \textbf{92.2} & 26.3 & 10.2 & \textbf{7.0} & 2.32 & \textbf{3.49} \\
Baseline Reseed & 30 & 25 & 3.0--6.0 & 79.3 & 90.6 & \textbf{99.2} & 12.6 & 7.0 & \textbf{0.8} & 2.33 & \textbf{3.53} \\
Very Dense & 80 & 25 & 3.0--6.0 & 34.5 & 87.5 & \textbf{93.8} & 28.8 & 10.9 & \textbf{6.2} & 2.32 & \textbf{3.43} \\
Very Sparse & 5 & 25 & 3.0--6.0 & 94.6 & 100.0 & 100.0 & 2.2 & 0.0 & 0.0 & 2.42 & \textbf{3.56} \\
Pure Spheres & 30 & 25 & 3.0--6.0 & 78.8 & 91.4 & \textbf{99.2} & 11.0 & 7.0 & \textbf{0.8} & 2.36 & \textbf{3.54} \\
Pure Cubes & 30 & 25 & 3.0--6.0 & 82.0 & 92.2 & \textbf{99.2} & 6.3 & 6.2 & \textbf{0.8} & 2.33 & \textbf{3.58} \\
Tight Arena & 12 & 10 & 3.0--5.0 & 72.5 & 89.1 & \textbf{100.0} & 6.3 & 10.2 & \textbf{0.0} & 1.93 & \textbf{2.87} \\
Ultra Fast & 30 & 25 & 7.0--10.0 & 26.8 & 90.6 & \textbf{94.5} & 72.0 & 9.4 & \textbf{5.5} & 2.65 & \textbf{5.09} \\
Ultra Slow & 30 & 25 & 0.5--1.5 & 1.4 & 87.5 & \textbf{98.4} & 2.7 & 4.7 & \textbf{1.6} & \textbf{1.02} & 0.93 \\
Tight Cluster & 50 & 25 & 3.0--6.0 & 36.1 & 87.5 & \textbf{89.1} & 38.1 & \textbf{8.6} & 10.9 & 2.33 & \textbf{3.45} \\
\midrule
\textbf{mean} & & & & 61.8 & 90.3 & \textbf{97.1} & 18.2 & 8.2 & \textbf{2.9} & 2.24 & \textbf{3.39} \\
\bottomrule
\end{tabular}
\end{table*}

\subsection{Controlling for commanded speed}
\label{sec:matched}

To separate the proximity term from the training speed band, we re-ran a subset of the suite with the commanded band forced to $\mathcal{U}[2,4]$~m/s for both policies on the same scenes and seeds, using OA-only rather than the full-bundle ClearNav. The full bundle was not re-run under matched velocity, so this controls the confound for $\Loa$ alone, not the bundle. Since $\mathcal{L}_{\mathrm{slip}}$ and the look-ahead yaw reference are not speed-band-dependent in the same way, we take what it establishes for $\Loa$ as a lower bound on the bundle.

Table~\ref{tab:matchedvel} shows success improving from $90.6\%$ to $97.2\%$ and collisions falling from $8.4\%$ to $2.6\%$, a $70\%$ relative reduction, over 512 paired episodes on the four re-run scenarios. The direction and magnitude of Table~\ref{tab:static} survive. The two policies still differ in speed under a matched command, $2.73$ vs $2.04$~m/s, so the control is partial, but it rules out the strong confound: with OA-only still faster, it collides less often, which a speed-only explanation does not predict.

\begin{table}[!htbp]
\vspace*{3pt}
\centering
\caption{Matched-velocity control, run on OA-only rather than the full-bundle ClearNav. Both policies are
commanded the same $v^\star \sim \mathcal{U}[2,4]$~m/s band, removing the
commanded-speed difference between the two training configurations. 128
episodes per scenario per policy. The full-bundle ClearNav was not re-run under
matched velocity, so this controls the confound for the proximity term alone,
not for the bundle; see Sec.~\ref{sec:matched}.}
\label{tab:matchedvel}
\small
\setlength{\tabcolsep}{5pt}
\resizebox{\columnwidth}{!}{%
\begin{tabular}{l rr rr rr}
\toprule
& \multicolumn{2}{c}{Success (\%)} & \multicolumn{2}{c}{Collision (\%)} & \multicolumn{2}{c}{Speed (m/s)} \\
\cmidrule(lr){2-3}\cmidrule(lr){4-5}\cmidrule(lr){6-7}
Scenario & B & OA-only & B & OA-only & B & OA-only \\
\midrule
Baseline & 89.1 & \textbf{100.0} & 9.4 & \textbf{0.0} & 2.03 & 2.71 \\
Sparse & 98.4 & 98.4 & 1.6 & 1.6 & 2.09 & 2.74 \\
Dense & 87.5 & \textbf{96.1} & 11.7 & \textbf{3.1} & 2.03 & 2.73 \\
Very Dense & 87.5 & \textbf{94.5} & 10.9 & \textbf{5.5} & 2.01 & 2.75 \\
\midrule
\textbf{mean} & 90.6 & \textbf{97.2} & 8.4 & \textbf{2.6} & 2.04 & 2.73 \\
\bottomrule
\end{tabular}}
\end{table}

\subsection{Moving obstacles}
\label{sec:moving}

$\Loa$ is computed from an instantaneous clearance without modeling obstacle motion, and neither policy saw a moving obstacle in training, so this suite probes transfer on both counts. Obstacles are relocated each control step on trajectories seeded identically for both policies. Eleven scenarios cover linear drift, crossing traffic, orbits about the goal, oscillating pillars, and a dense fast swarm, which we showed in the supplementary video. Five more mix movers with a fixed static layout. Each scenario is 8 scenes $\times$ 3 rollouts $=$ 24 paired episodes per policy with a 600-step limit. We report the suite with obstacle--obstacle separation enforced and, secondarily, without it. The correction is deterministic and pairing is preserved in both variants.

Table~\ref{tab:moving} shows the results. With separation enforced, mean success improves from $64.0\%$ to $73.9\%$ and collisions fall from $36.0\%$ to $26.1\%$, winning 7 scenarios, losing 3 and tying 1, but this does not reach significance at $n = 11$, $p = 0.084$. Without separation, the effect is of similar magnitude, $68.6\% \rightarrow 76.1\%$ with $p = 0.19$, and in the mixed scenarios ClearNav leads $80.0\%$ to $71.7\%$, again not significant. We interpret this as no evidence that ClearNav hurts on moving obstacles and a weak, consistent-in-sign suggestion that it helps, with 24 episodes per scenario too few to establish it. The mechanism is also plausibly limited: it slows near a wall, but if drone approaches to the obstacle at $5 \si m/s$, the safe action may be to accelerate away.

\begin{table}[!htbp]
\vspace*{3pt}
\centering
\caption{Moving-obstacle suite. Obstacles are relocated at the control rate
on trajectories seeded identically for both policies. 24 episodes per
scenario per policy. Primary columns enforce obstacle--obstacle separation; the
right-hand pair repeats without it. $v_o$ is a range for orbiters
($|\omega| r$, $|\omega|\in[0.3,0.8]$~rad/s) and oscillators (speed varies
within an episode up to the quoted peak).}
\label{tab:moving}
\footnotesize
\setlength{\tabcolsep}{2.5pt}
\renewcommand{\arraystretch}{0.85}
\resizebox{\columnwidth}{!}{%
\begin{tabular}{l cc rr rr rr}
\toprule
& & & \multicolumn{2}{c}{Success (\%)} & \multicolumn{2}{c}{Collision (\%)} & \multicolumn{2}{c}{Succ., no sep.} \\
\cmidrule(lr){4-5}\cmidrule(lr){6-7}\cmidrule(lr){8-9}
Scenario & $|\mathcal{O}|$ & $v_o$ & B & Ours & B & Ours & B & Ours \\
\midrule
Slow Drift & 10 & 0.5 & 95.8 & \textbf{100.0} & 4.2 & \textbf{0.0} & 91.7 & 95.8 \\
Crossing & 10 & 2.0 & 70.8 & 70.8 & 29.2 & 29.2 & \textbf{79.2} & 70.8 \\
Orbiting & 12 & 0.9--8.8 & \textbf{45.8} & 41.7 & \textbf{54.2} & 58.3 & 45.8 & \textbf{62.5} \\
Oscillating & 12 & 0--4.8 & 66.7 & \textbf{95.8} & 33.3 & \textbf{4.2} & 62.5 & \textbf{100.0} \\
Dense Swarm & 18 & 1.3 & 66.7 & \textbf{83.3} & 33.3 & \textbf{16.7} & 79.2 & \textbf{91.7} \\
Crossing v1 & 10 & 1.0 & 66.7 & \textbf{100.0} & 33.3 & \textbf{0.0} & 83.3 & \textbf{87.5} \\
Crossing v3 & 10 & 3.0 & 54.2 & \textbf{66.7} & 45.8 & \textbf{33.3} & 50.0 & \textbf{70.8} \\
Crossing v4 & 10 & 4.0 & \textbf{66.7} & 58.3 & \textbf{33.3} & 41.7 & \textbf{70.8} & 66.7 \\
Crossing v5 & 10 & 5.0 & 29.2 & \textbf{58.3} & 70.8 & \textbf{41.7} & 37.5 & \textbf{58.3} \\
Swarm Fast & 18 & 2.5 & \textbf{58.3} & 50.0 & \textbf{41.7} & 50.0 & \textbf{83.3} & 62.5 \\
Drift Fast & 10 & 2.0 & 83.3 & \textbf{87.5} & 16.7 & \textbf{12.5} & 70.8 & 70.8 \\
\midrule
\textbf{mean} & & & 64.0 & \textbf{73.9} & 36.0 & \textbf{26.1} & 68.6 & \textbf{76.1} \\
\midrule
\multicolumn{9}{l}{\emph{mixed: static obstacles $+$ movers}} \\
Static Drift & 6$+$8 & 0.8 & 95.8 & 95.8 & 4.2 & 4.2 & -- & -- \\
Static Crossing & 6$+$8 & 2.0 & 87.5 & 87.5 & 12.5 & 12.5 & -- & -- \\
Pillars Swarm & 10$+$10 & 1.5 & 66.7 & \textbf{70.8} & 33.3 & \textbf{29.2} & -- & -- \\
Dense Mixed & 12$+$12 & 1.5 & 62.5 & 62.5 & 37.5 & 37.5 & -- & -- \\
Static Orbit & 8$+$8 & 0.9--8.8 & 45.8 & \textbf{83.3} & 54.2 & \textbf{16.7} & -- & -- \\
\midrule
\textbf{mean} & & & 71.7 & \textbf{80.0} & 28.3 & \textbf{20.0} & -- & -- \\
\bottomrule
\end{tabular}}
\end{table}

The crossing-traffic sweep in Table~\ref{tab:speedsweep} isolates obstacle speed. The sweep is run on OA-only to characterize the proximity term's transfer to dynamic obstacles. Both policies degrade sharply with crossing speed, from $80$--$95\%$ success at 1~m/s to $33$--$38\%$ at 5~m/s, and the ordering between them is not stable. Neither policy handles fast dynamic obstacles, unsurprising since neither saw one in training, and the proximity term does not change this.

\begin{table}[!htbp]
\vspace*{3pt}
\centering
\caption{Crossing-traffic speed sweep, run on OA-only.
All five columns share obstacle seed 102, so the obstacle
field and headings are identical and only the speed magnitude changes. Rates in
\%.}
\label{tab:speedsweep}
\small
\begin{tabular}{l ccccc}
\toprule
& \multicolumn{5}{c}{obstacle speed $v_o$ (m/s)} \\
\cmidrule(lr){2-6}
& 1 & 2 & 3 & 4 & 5 \\
\midrule
B success & 83.3 & 79.2 & 50.0 & 70.8 & 37.5 \\
OA-only success & 95.8 & 54.2 & 58.3 & 70.8 & 33.3 \\
B collision & 16.7 & 20.8 & 50.0 & 29.2 & 62.5 \\
OA-only collision & 4.2 & 45.8 & 41.7 & 29.2 & 66.7 \\
\bottomrule
\end{tabular}
\end{table}

\subsection{Cross-framework comparison}
\label{sec:crossframework}

The proximity cost \eqref{eq:loa} is a direction-agnostic term we introduce (Sec.~\ref{sec:newloss}). For context, the \emph{DA} columns of Table~\ref{tab:static} evaluate the DiffAero reference policy~\cite{zhangDiffAeroGPUAcceleratedDifferentiable2025} on the same 18 scenario specifications in the DiffAero simulator. On the 18 ClearNav scenarios, DA's mean success is $61.8\%$ against $90.3\%$ for B and $97.1\%$ for ClearNav. This is not a model comparison. The stacks differ in renderer, depth resolution, dynamics, termination criteria, and episode limits, so the numbers benchmark whole stacks, not the reward term.

\subsection{Real world flight}
\label{sec:realworld}

We also fly the ClearNav policy on a physical quadrotor to confirm that the simulator-trained behavior transfers to real hardware, with the policy running onboard on a Jetson Orin NX (Fig.~\ref{fig:realworld}). The scene holds four obstacles of different shapes along the path to the goal. The first is narrow, so the drone moves left to avoid it, then once it has looked at the obstacle it tries to go below it. Since there is ground below it does not move further and it keeps its altitude low while trying to yaw around the second obstacle, which is a wider variety, so the drone intentionally yaws to the right to clear it before continuing the rest of the trajectory to the goal. The baseline policy was not successful under the same conditions. We also tested the baseline policy in the same real-world scenario, but it flew too broadly and grazed the nets enclosing the flying area, or crashed directly into the net. We therefore conclude that ClearNav successfully transferred to real-world scenario.

\section{Conclusion}\label{sec:conclusion}

We presented ClearNav, a navigation policy for autonomous quadrotor flight in cluttered and confined spaces, where a small platform carries only a single forward-facing depth camera and proprioception. Three reward terms proposed in this paper on penalizing high speed near obstacles regardless of direction, turning the heading toward the geodesic gradient sampled ahead of the vehicle, and making sideways flight expensive makes ClearNav a safer policy than the baseline policies trained in differentiable physics simulator. Every behavior the objective penalizes is observable at inference. On the training distribution the proximity term costs nothing measurable. On 18 held-out static scenarios, ClearNav raises success from $90.3\%$ to $97.1\%$ and cuts collisions by $65\%$, winning 17 of 18 while flying faster. On matched velocity control isolation the proximity term preserves a $70\%$ collision reduction. On moving obstacles the effect points the same way but does not reach significance. We also flew the ClearNav policy on a physical quadrotor with the policy running onboard, and it navigated through a cluttered scene safely that the baseline could not, confirming that the behavior transfers to real hardware. One limitation is privileged supervision. Since the collision and proximity losses use an exact omnidirectional clearance query while the policy observes only a single $89^\circ$ depth camera, so a learned maneuver need not be supported by an observation, a gap we plan to address in future work.

\bibliographystyle{IEEEtran}

\bibliography{mybib}

\end{document}